\documentclass[runningheads]{llncs}
\usepackage{graphicx}
\usepackage{amsmath}
\usepackage{amssymb}
\usepackage{hyperref}
\usepackage{booktabs}
\usepackage{xspace}
\usepackage{xcolor}
\usepackage{array}

\usepackage{multirow}
\usepackage{tikz}
\usepackage{booktabs} 
\usepackage{multicol}
\usepackage{subfigure}
\usepackage{threeparttable}
\usepackage{mathrsfs}
\usepackage{makecell}

\makeatletter
\newcommand{\thickhline}{%
\noalign {\ifnum 0=`}\fi \hrule height 1pt
\futurelet \reserved@a \@xhline
}
\makeatother

\usepackage[table]{xcolor}       
\definecolor{mygray}{gray}{.9}  
\PassOptionsToPackage{numbers}{natbib}
\definecolor{url}{RGB}{0,73,147}
\definecolor{mypink}{HTML}{bc4749}

\usepackage{xspace}
\usepackage{bbding}
\usepackage{wrapfig}
\makeatletter
\DeclareRobustCommand\onedot{\futurelet\@let@token\@onedot}
\def\@onedot{\ifx\@let@token.\else.\null\fi\xspace}

\makeatletter
\DeclareRobustCommand\onedot{\futurelet\@let@token\@onedot}
\def\@onedot{\ifx\@let@token.\else.\null\fi\xspace}
\def\eg{\emph{e.g}\onedot}

\def\etal{\emph{et al}\onedot}
\makeatother

\begin{document}

\title{A Multi-Modal Framework with Cross-Subject Pseudo-Labeling and Semantic Alignment for Micro-Gesture Recognition}

\titlerunning{Multi-Modal Framework for Micro-Gesture Recognition} 

\author{Haoran Zhang*\inst{1} \and Haokun Zhang*\inst{2} \and Pengyu Liu\inst{1} \and Yujia Zhang\inst{1} \and Weibao Xue\inst{1} \and Yanbin Hao$^{\dagger}$\inst{1} }

\authorrunning{H. Zhang\etal} 

\institute{$^1$ School of Computer Science and Information Engineering, Hefei University of Technology (HFUT), Hefei, China \\
$^2$ School of Computer Science, University of Auckland(UOA), Auckland, New Zealand \\
\email{modeliqi9@gmail.com, hzha314@aucklanduni.ac.nz, lpynow@gmail.com, 18226601579@163.com, xuewbb@gmail.com, haoyanbin@hotmail.com}} 

\footnotetext[1]{*: Co-first author.}
\footnotetext[2]{$^{\dagger}$: Corresponding author.}

\maketitle

\begin{abstract}
Micro-gestures (MGs) are spontaneous and subtle body movements that frequently convey hidden human emotions. Recognizing MGs in untrimmed videos remains highly challenging due to their extremely low signal-to-noise ratio, severe long-tailed class distribution, and the inherent domain shift encountered in cross-subject evaluation scenarios. In this paper, we propose a comprehensive multi-modal framework for Track 1 of the 4th MiGA-IJCAI Challenge. To capture fine-grained representations, we design a saliency-guided multi-modal extraction pipeline integrating 68-keypoint skeleton joint coordinates, 3D heatmap volumes, and high-resolution RGB visual features. We introduce a gentle square-root smoothed weighting mechanism paired with an Orthogonal Semantic Embedding Loss to protect tail classes without compromising overall recognition capabilities. 
% 对应
More importantly, to bridge the cross-subject generalization gap, we propose a Cross-Modal Pseudo-Labeling (CMPL) strategy for unsupervised domain adaptation, which significantly boosts single-modal robustness. A temperature-scaled soft-voting mechanism is finally utilized to alleviate overconfidence during late fusion. Extensive experiments demonstrate that our framework achieves a competitive F1-score of 68.13\%, securing the 4th place.

\keywords{Micro-gesture Recognition \and Multi-modal Fusion \and Unsupervised Domain Adaptation \and Long-tailed Learning.}
\end{abstract}

\section{Introduction}

Micro-gestures (MGs) and micro-actions are spontaneous, subtle body movements (\eg, Rubbing the nose) that reliably indicate hidden human emotions~\cite{ekman}. Recognizing them automatically holds substantial practical significance for affective computing and clinical psychology~\cite{imigue,smg,liu2025survey,casme2}. 
% Micro-gesture recognition (MGR)
However, Micro-gesture recognition (MGR) in untrimmed videos remains a formidable task, serving as the core objective of Track 1 in the 4th MiGA-IJCAI Challenge\protect\footnotemark[1]. 
% 一般不这么讲：可以写MGR的动作和背景区分小，动作幅度低，信噪比低，是MGR的难点，cross subject一般不单独写出来
The primary difficulties of MGR stem from the intrinsically low action amplitudes and the minimal visual distinction between subtle movements and complex backgrounds, which collectively result in a severely low signal-to-noise ratio and make fine-grained recognition highly challenging.

\footnotetext[1]{The Kaggle competition page: \href{https://www.kaggle.com/competitions/the-4th-ei-mi-ga-ijcai-challenge-track-1/leaderboard}{https://www.kaggle.com/competitions/the-4th-ei-mi-ga-ijcai-challenge-track-1/leaderboard}}

% 转折词用的不对，是两种并行的方法，看一下别人的论文这里怎么写的
To address these challenges, existing methodologies primarily rely on two parallel streams: skeleton-based and RGB-based approaches. Skeleton-based models \cite{stgcn,ctrgcn,shiftgcn} explicitly capture physical body topology but are sensitive to pose estimation errors and lack visual textures. On the other hand, RGB-based networks \cite{i3d,videoswin,timesformer} encode rich spatio-temporal contexts but often struggle to isolate low-amplitude MGs from complex background noise. While multi-modal fusion attempts to integrate both \cite{mmtm}, conventional paradigms often suffer from modality dominance.
% 可以写虽然以前的这些工作都针对MGR的信噪比低做出了一些改进，但是我们发现，长尾分布、cross subject是两个关键的islight，会在哪些方面影响性能
Although previous efforts have made improvements in mitigating the low signal-to-noise ratio of MGR, we observe that the extreme long-tailed distribution and cross-subject evaluation are two critical bottlenecks restricting the ultimate performance. Specifically, the severe long-tailed nature inherently biases the optimization process towards majority classes, resulting in poor discrimination of rare micro-gestures. Furthermore, the strict cross-subject protocol introduces significant domain shifts, leading to sharp performance degradation when the learned models are generalized to unseen individuals.

% 写的太模糊了，RGB裁剪有做吗
% 多模态微手势识别框架
To overcome these bottlenecks, we propose a robust Multi-Modal Micro-gesture Framework that effectively integrates RGB and skeleton modalities to achieve noise suppression and feature complementarity. Specifically, rather than using raw full-frame videos, our vision branch performs saliency-guided cropping based on skeleton bounding boxes to isolate human subjects, thereby strictly filtering out background noise. These cropped high-resolution RGB clips are then processed by Video Swin Transformer \cite{videoswin} and R(2+1)D \cite{r2plus1d} to encode rich visual textures. In parallel, the skeleton branch leverages 68-keypoint sequences to capture physical topologies, utilizing a Decoupled Spatial-Temporal CNN for joint coordinates and a 3D-ResNet for rendered heatmaps \cite{posec3d}. To combat the extreme long-tailed distribution, we employ a square-root smoothed sampling strategy coupled with an Orthogonal Semantic Embedding Loss \cite{li2023joint} to explicitly enlarge inter-class margins. Furthermore, to bridge the domain gap caused by unseen subjects, we design an iterative Cross-Modal Pseudo-Labeling (CMPL) \cite{pseudolabel,fixmatch}strategy, leveraging high-confidence consensus on the test set to formulate a super-dataset for unsupervised domain adaptation. Finally, multi-modal predictions are integrated through a temperature-scaled soft-voting fusion mechanism \cite{guo2017calibration} to alleviate model overconfidence and guarantee optimal prediction reliability.

% 和上一段合并起来，用一段内容大概描述怎么做的
% 图上画的其实是两个分支
% 温度投票加引用

\section{Related Work}

\subsection{Micro-Gestures Dataset}

Recognizing hidden emotions requires meticulously annotated and challenging benchmarks. The iMiGUE \cite{imigue} dataset was specifically introduced as an identity-free dataset aimed at understanding suppressed emotions through spontaneous MGs. Unlike traditional gesture datasets that focus on illustrative macro-movements, iMiGUE captures involuntary gestures elicited during highly stressful scenarios (\eg, post-match press conferences). It comprises 359 videos from 72 subjects across 28 countries, posing a significant challenge for cross-subject generalization.

To comprehensively analyze subtle human behaviors, the research community has progressively developed a series of specialized benchmarks. For instance, the MA-52 dataset \cite{guo2024benchmarking} and related studies \cite{gu2025motion} were established to evaluate full-body spontaneous micro-actions during psychological interviews. Subsequently, the MMA-52 dataset \cite{li2025mmad} introduced multi-label annotations to support the complex task of concurrent micro-action detection in untrimmed videos. More recently, MA-Bench \cite{li2026bench} was proposed to provide a finer-grained hierarchical evaluation platform. Driven by these rich datasets, the Micro-Action Analysis Challenges \cite{guo2024mac,li2025mac} have been hosted consecutively, successfully shifting the evaluation paradigms from isolated classification to continuous online detection tasks \cite{filntisis2020emotion,liu2024micro,liu2025online}. 

\subsection{Multi-Modal Action Recognition}

% 连接词不对，写的太多了
Action recognition has significantly progressed with the evolution of deep learning architectures. For RGB modalities, 3D CNNs such as I3D \cite{i3d}, SlowFast \cite{slowfast}, and R(2+1)D \cite{r2plus1d} laid the foundation for video modeling. Recently, Vision Transformers like VideoMAE \cite{videomae} , Video Swin Transformer \cite{videoswin}, TimeSformer \cite{timesformer}, and UniFormer \cite{uniformer} have shown superior capacity in capturing long-range visual dependencies. Furthermore, advanced contextual modeling and attention mechanisms \cite{hao2022group,hao2022attention} have been developed to efficiently capture spatial-temporal correlations in videos. Conversely, skeleton-based methods focus purely on the physical topology of human bodies, which has proven highly effective in general pose action recognition \cite{li2023data}. Graph Convolutional Networks (GCNs), such as ST-GCN \cite{stgcn} and CTR-GCN \cite{ctrgcn}, Shift-GCN \cite{shiftgcn}, and HD-GCN \cite{hdgcn}, have been widely adopted by recent micro-gesture recognition studies \cite{li2023joint,chen2024prototype,gu2025mm,li2025prototypical} to precisely extract structural motion features.

% 多加几个方法
However, purely coordinate-based methods are frequently sensitive to topological noise and missing joints, which are prevalent in in-the-wild videos. To address this, Duan \etal proposed PoseC3D \cite{posec3d}, rendering skeleton sequences into 3D pseudo-heatmaps. Meanwhile, to fully leverage heterogeneous features, recent works have also explored cross-modal feature enhancement and contrastive alignment \cite{shang2025cross} to boost micro-gesture recognition. Following the success of these mainstream paradigms, our framework explicitly leverages Video Swin Transformer and R(2+1)D for high-resolution RGB visual textures, alongside PoseC3D and a Decoupled ST-CNN for robust structural joint cues, formulating a holistic multi-modal representation specifically tailored for fine-grained micro-gesture analysis.

\subsection{Cross-Modal Pseudo-Labeling and Long-Tailed Learning }

% 引用，以前的方法不好，我们的好
{Real-world datasets inherently exhibit long-tailed class distributions. Traditional methods to calibrate such imbalance fall into re-sampling,  re-weighting (\eg, Focal Loss \cite{focalloss}), and decoupled representation learning \cite{decoupling,bbn}. While aggressive over-sampling is frequently utilized, we empirically observed it leads to severe mode collapse in extreme micro-gesture classification. Beyond standard calibration, recent works \cite{li2023joint} utilize natural language word vectors to guide the latent feature space. Inspired by this, we integrate an Orthogonal Semantic Embedding Loss. Furthermore, cross-subject generalization requires aligning feature distributions across domains , often leveraging adversarial training \cite{dann} or source hypothesis transfer \cite{shot}.Pseudo-labeling \cite{pseudolabel,fixmatch} is a powerful technique to handle this. In our pipeline, we propose an iterative Cross-Modal Pseudo-Labeling (CMPL) strategy to significantly boost cross-subject generalization performance across diverse unseen domains.}

\section{Methodology}

\begin{figure}[htbp]
    \centering
    \includegraphics[width=\textwidth]{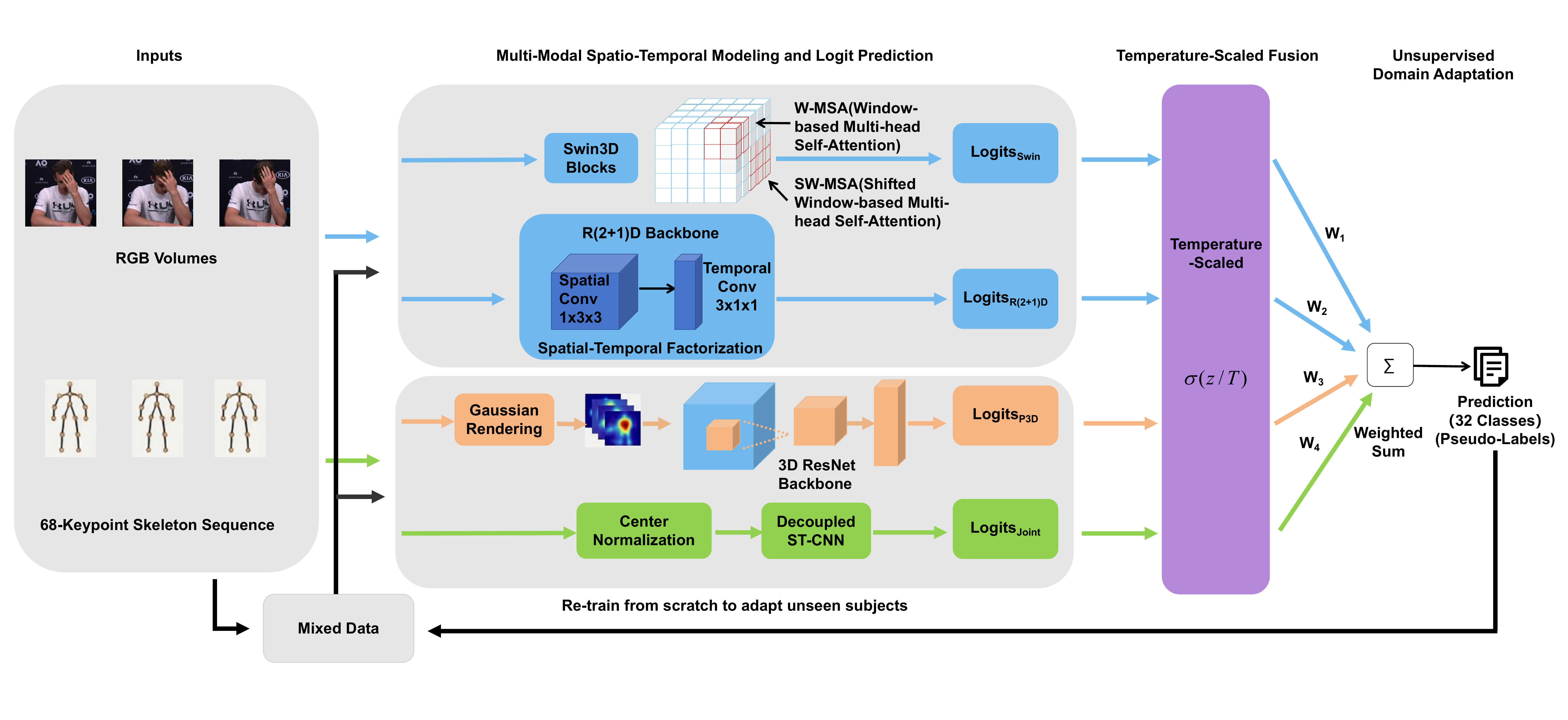} \caption{The overall architecture of our proposed framework. First, raw RGB volumes and 68-keypoint skeletons are fed into four parallel branches (Swin3D, R(2+1)D, PoseC3D, and Decoupled ST-CNN) for multi-modal spatio-temporal modeling. Second, the generated logits are calibrated and aggregated via a Temperature-Scaled Fusion ($\sigma(z/T)$) to mitigate individual model overconfidence. Finally, an iterative pseudo-labeling loop forms a Data super-dataset for Unsupervised Domain Adaptation, enabling models to re-train from scratch and seamlessly adapt to unseen subjects}
    \label{fig:architecture}
\end{figure}

% motivation不清楚，然后要对整个main figure做一个描述
To effectively recognize micro-gestures, relying on a single modality is fundamentally insufficient due to the extremely low signal-to-noise ratio and subtle motion amplitudes. Our motivation is to construct a holistic representation that synergizes the explicit physical topology from skeletons, the spatio-temporal continuity from 3D heatmaps, and the fine-grained visual textures from RGB clips. Furthermore, to inherently address the extreme long-tailed class distribution and cross-subject domain shifts, we tightly integrate semantic regularization and unsupervised domain adaptation into our pipeline. The overall architecture of our proposed framework is explicitly illustrated in Fig. \ref{fig:architecture}. 

% 【图表 1：整体架构图】

\subsection{Saliency-Guided Multi-Modal Encoding}

% micro expression?
\textbf{Skeleton Joint Branch:} Traditional 35-keypoint skeleton graphs fail to capture crucial facial micro-movements, such as biting lips or subtle jaw shifts. Therefore, we utilize the OpenPose toolkit \cite{openpose} to extract $V=68$ keypoints from the raw videos, deliberately incorporating 19 precise facial landmarks along with hand and body joints. Let $J \in \mathbb{R}^{C_{in} \times T \times V}$ denote the joint coordinates, where $C_{in}=3$ represents $(x, y, \text{confidence})$ and $T$ is the temporal length. To eliminate absolute spatial variations, we apply Center Normalization by subtracting the center joint coordinates before feeding them into the network. We process this sequence using a Decoupled Spatial-Temporal CNN, which independently applies $(1 \times 3)$ spatial convolutions and $(5 \times 1)$ temporal convolutions to efficiently capture the fine-grained spatial displacements of the continuous joint coordinate sequences with high structural fidelity.

\textbf{3D Heatmap Branch (PoseC3D):} 
% motivation 不对
Purely coordinate-based models are intrinsically vulnerable to coordinate jittering and tracking failures, which can easily overwhelm the low-amplitude motion signals of micro-gestures. To construct a more resilient structural representation, we adopt PoseC3D \cite{posec3d}.The 68 joint coordinates are rendered into a 3D heatmap volume $\mathcal{H} \in \mathbb{R}^{V \times T \times H \times W}$, where $V=68$ represents independent channels for each keypoint, and $H, W$ represent the spatial resolution. A 3D-ResNet50 is adopted as the backbone. By interpreting coordinates as Gaussian heatmaps, 3D convolutions effortlessly bridge short-term joint disappearances through spatial-temporal continuity. We also enable horizontal flipping during target generation to effectively augment the spatial receptive field during the network training phase.

\textbf{Saliency-Guided Vision Branch:} Skeletons fundamentally discard appearance context, making it impossible to capture subtle facial deformations (\eg, biting lips) or fine-grained hand-body interactions (\eg, adjusting a collar). To complement this structural deficiency, the RGB modality is indispensable for providing dense visual cues.  However, feeding entire raw frames introduces severe background noise. Thus, we introduce a Saliency-Guided Cropping module. We dynamically crop the Region of Interest (ROI) bounding boxes guided by the extreme boundaries of the skeleton coordinates (with a 20\% safety margin). These cropped subjects are magnified to $224 \times 224$ high resolution. The visual features are then extracted using Video Swin Transformer (Swin3D) \cite{videoswin} and R(2+1)D \cite{r2plus1d} pre-trained on Kinetics-400 to extract rich spatial-temporal visual textures from the cropped regions.

\subsection{Rethinking Long-Tailed Calibration in Extreme Imbalance}

\begin{figure}[t] 
    \centering
    \includegraphics[width=0.95\textwidth]{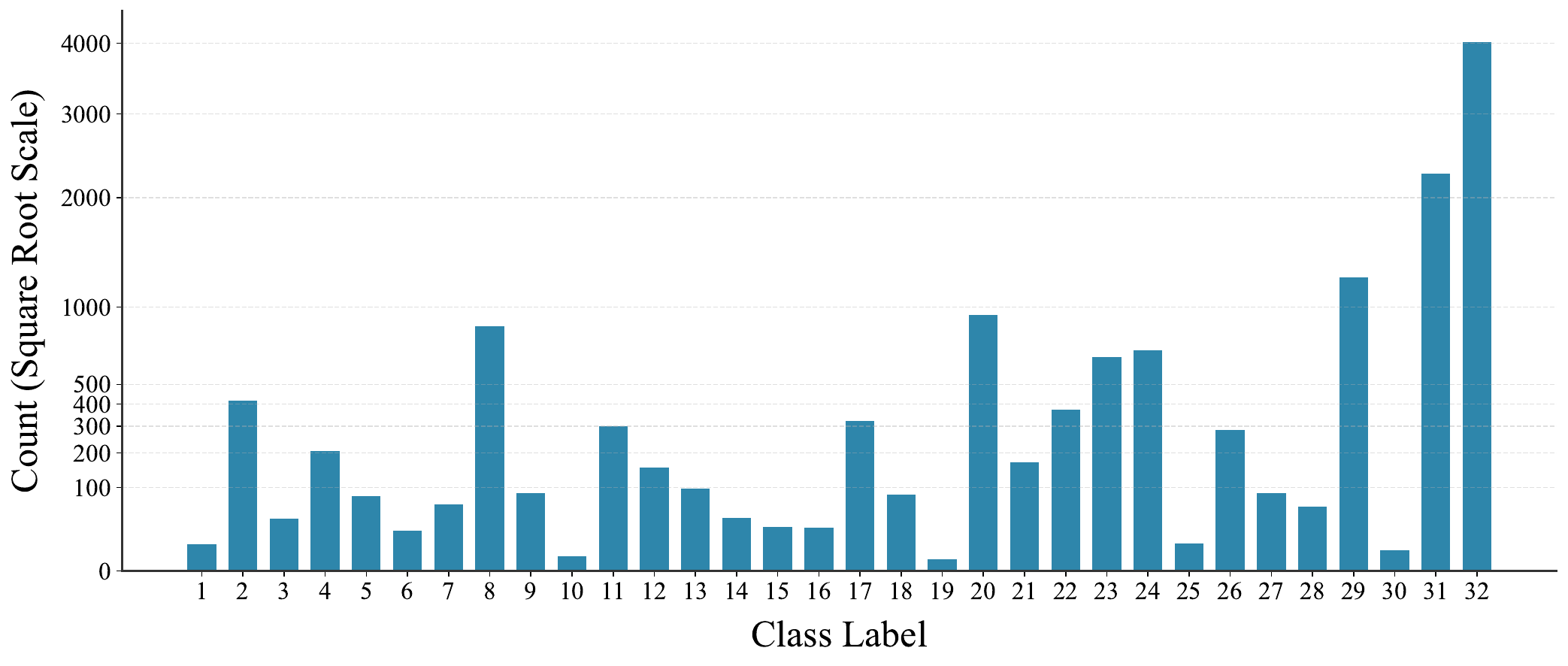} 
    \caption{Illustration of the extreme long-tailed class distribution in the iMiGUE dataset. The severe imbalance ratio ($> 2000:1$) often leads to catastrophic mode collapse if traditional aggressive re-sampling techniques are applied.}
    \label{fig:distribution} 
\end{figure}

% 写的不对
The iMiGUE dataset inherently exhibits a severe long-tailed class distribution. As explicitly illustrated in Fig. \ref{fig:distribution}, a few head classes (particularly the non-micro-gesture background category) naturally dominate the dataset, while the vast majority of target micro-gesture categories suffer from extremely scarce training instances. To address this severe data imbalance issue, we propose a soft regularization approach:

% 该引用的引用
\textbf{1) Square-Root Smoothed Weighting:} To prevent gradient explosion for rare classes---a common issue when directly applying inverse-frequency weighting \cite{cbloss}---we calculate the occurrence frequency $N_c$ for each class $c$. We define a smoothed penalty weight $w_c = 1 / \sqrt{N_c + 1}$. This weight is effectively integrated into the Focal Loss \cite{focalloss}, gently encouraging the model to attend to minority classes without memorizing specific instances or overfitting to background noise, which is a crucial principle for robust long-tailed learning \cite{ldam}.

\begin{figure}[t] 
    \centering
    \includegraphics[width=0.85\textwidth]{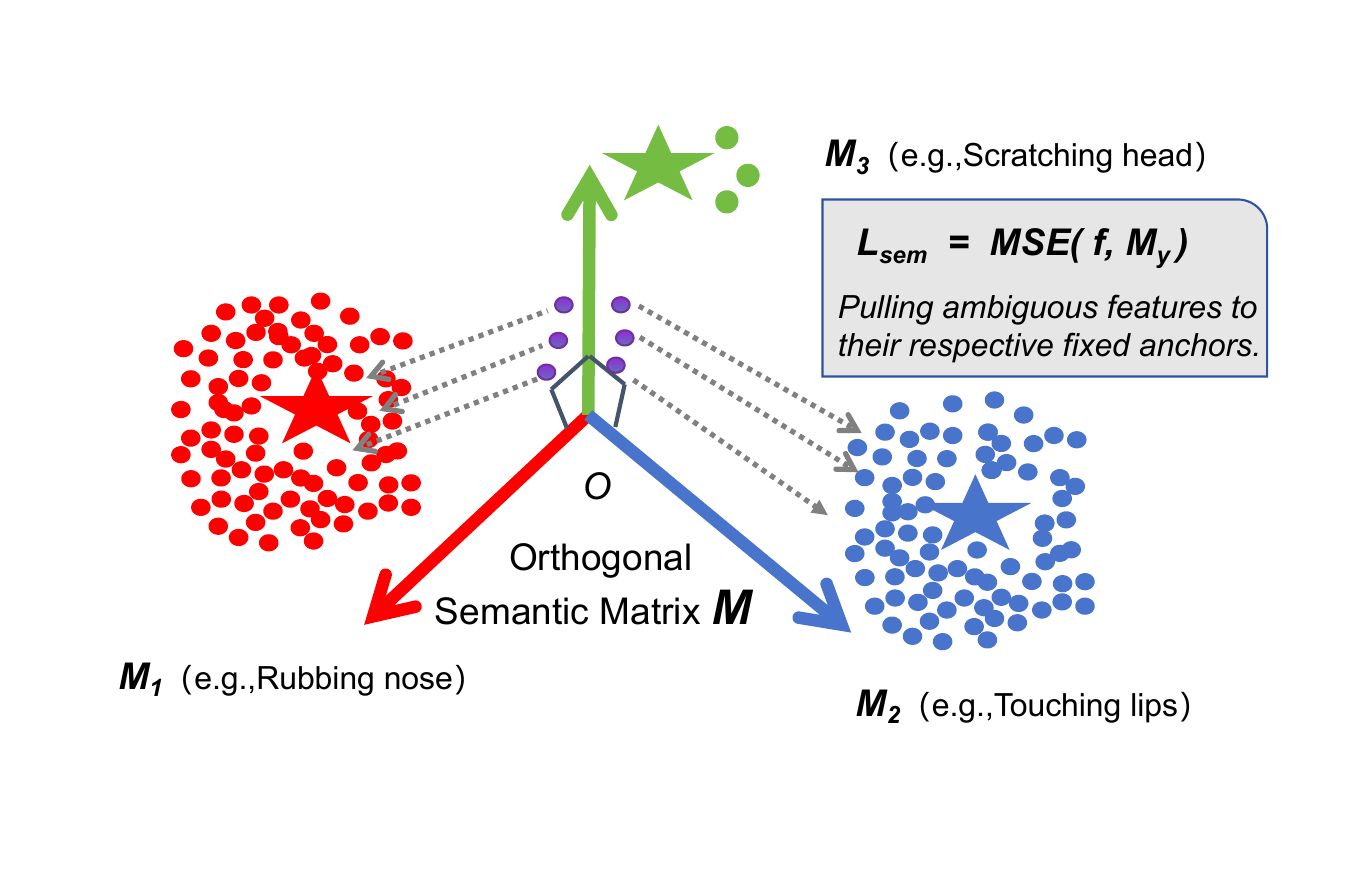} 
    \caption{A conceptual diagram illustrating the Orthogonal Semantic Embedding Loss. This mechanism explicitly pulls ambiguous visual features toward their respective fixed orthogonal anchors, effectively preventing minority tail classes from being swallowed by dominant majority head classes within the latent feature space.}
    \label{fig:semantic}
\end{figure}

\textbf{2) Orthogonal Semantic Embedding Loss:} Due to the high visual similarity among micro-gestures, we introduce an auxiliary semantic loss, inspired by recent advancements in visual-semantic alignment \cite{li2023joint}. Let $f \in \mathbb{R}^d$ be the extracted visual feature before the classifier. We define a non-trainable semantic matrix $M \in \mathbb{R}^{K \times d}$ representing $K=32$ action classes with orthogonal initialization (generated via QR decomposition of a random matrix). While fixed anchors might impose strict constraints, they act as a powerful structural prior (analogous to prototype learning \cite{chen2024prototype,li2025prototypical}) to prevent the model from overfitting to the background noise of tail classes. As shown in Fig. \ref{fig:semantic}, the semantic alignment loss is formulated as the Mean Squared Error (MSE) between the feature and the target semantic anchor:
\begin{equation}
    \mathcal{L}_{sem} = \text{MSE}(f, M_y) = \frac{1}{d} \sum_{i=1}^{d} (f_i - M_{y, i})^2
\end{equation}
where $y$ is the ground-truth label. This explicit constraint forces the network to enlarge the inter-class distance among highly confusing micro-gestures within the latent semantic embedding space \cite{li2023joint}.

% 【图表 2：长尾分布柱状图 (独立排版)】

% 【图表 3：正交语义原理图 (独立排版)】

\subsection{Cross-Modal Pseudo-Labeling (CMPL)}

To bridge the domain gap inherent in the cross-subject testing protocol, we introduce a powerful UDA strategy. We first aggregate our baseline multi-modal models to generate high-confidence predictions $\hat{Y}_{test}$ on the unlabeled test set $\mathcal{D}_{test}$. Specifically, we select target samples whose maximum softmax probability across all 32 action categories exceeds a stringent confidence threshold $\alpha$. Based on empirical evaluations, we set $\alpha = 0.65$ to guarantee the reliability of the generated annotations. We then treat these filtered predictions as pseudo-labels and inject them into the training set, forming an augmented training set $\mathcal{D}_{aug} = \mathcal{D}_{train} \cup \mathcal{D}_{val} \cup \mathcal{D}_{test}(\hat{Y}_{test})$.

By performing a single round of retraining from scratch on this super-dataset, the models implicitly learn the idiosyncratic behavioral nuances and fine-grained micro-action patterns of the unseen subjects while avoiding confirmation bias. This exposure drastically shifts the data distribution and lifts the single-modal accuracy ceiling without requiring any manual annotations or target domain labels.

\subsection{Temperature-Scaled Multi-Modal Fusion}

% 引用，引号要用`` ''
Late fusion through direct probability addition often suffers from the ``veto effect'', where an overconfident but incorrect model dominates the ensemble. To address this, we apply Temperature Scaling before the softmax operation to calibrate the logits $L_{m}$ from model $m$:
\begin{equation}
    P_m^{(c)} = \frac{\exp(L_m^{(c)} / \tau)}{\sum_{j=1}^{K} \exp(L_m^{(j)} / \tau)},
\end{equation}
where $\tau$ is the temperature parameter (set to 2.5 via grid search on the validation set). This technique softens sharp probability distributions, democratizing the voting process. 

The final fused probability is obtained via a weighted sum: 
\begin{equation}
    P_{final}^{(c)} = \sum_{m=1}^{M} W_m P_m^{(c)},
\end{equation}
where the modality weights $W_m$ are strictly determined through grid search based on validation performance, allowing fine-grained corrective cues from auxiliary models to influence the final prediction effectively and seamlessly mitigate the veto effect.

\section{Experiments}

\subsection{Dataset and Evaluation Metric}

The iMiGUE dataset \cite{imigue} is collected to evaluate hidden emotion understanding through micro-gestures. It comprises 32 classes of spontaneous micro-gestures. We utilize a strict cross-subject evaluation protocol, dividing the involved subjects into distinct training, validation, and testing groups without any overlapping identities. Furthermore, we employ the Top-1 Accuracy as our primary evaluation metric, which strictly corresponds to the Micro F1-Score within this multi-class setting.
% 我们使用了xxx

\subsection{Implementation Details}

% 这种写法很像实验报告，该引用的要引用
The \textbf{Skeleton Joint model} processes center-normalized 68-keypoint coordinate sequences, which are uniformly sampled to 64 frames. To optimize this network, we utilize the AdamW optimizer \cite{adamw} paired with a Cosine Annealing learning rate schedule \cite{cosine}. For the \textbf{PoseC3D model} \cite{posec3d}, 3D pseudo-heatmaps are rendered at a spatial resolution of $56 \times 56$ across 48 temporal frames. This branch is optimized using Stochastic Gradient Descent (SGD) for 100 epochs, employing a large batch size of 64 per GPU to ensure stable gradient convergence. Regarding the \textbf{RGB Vision models}, each cropped video clip is temporally sampled to yield 24 to 32 frames at a spatial resolution of $224 \times 224$. To further bolster the robustness of visual representations against spatial variances, we perform Test-Time Augmentation (TTA) during the inference phase by averaging the output logits from the original and horizontally flipped video sequences.

\subsection{Main Results}

Table \ref{tab:sota} illustrates the performance comparison with top-performing  methods from current and previous challenges on the iMiGUE cross-subject test set. Benefiting from the robust multi-modal feature encoding and the powerful cross-subject pseudo-labeling adaptation, our proposed ultimate ensemble achieves a highly competitive accuracy of 68.13\%,securing the 4th place in the challenge.

% 【图表 4：历届 SOTA 对比表格 (绝对防越界版)】
\begin{table}[htbp]
\centering
\caption{Top submissions of the past four MiGA Track-1 editions (2023--2026) on the iMiGUE test split, ranked by top-1 accuracy. Our submission (AIM) is highlighted in bold. Modalities: \textbf{J}=joint, \textbf{L}=limb, \textbf{R}=RGB, \textbf{T}=Taylor video, \textbf{F}=optical flow, \textbf{D}=depth video, \textbf{H}=heatmap.}
\label{tab:sota}
\renewcommand{\arraystretch}{1.2} 
\resizebox{\textwidth}{!}{%  <--- 核心魔法：强行把表格缩放到页面宽度
\begin{tabular}{c l c c c} 
\toprule
\rowcolor{mygray}
\textbf{Rank} & \textbf{Team} & \textbf{Core Methodology} & \textbf{Modality} & \textbf{Acc. (\%)} \\
\midrule

MiGA'23 1st & HFUT-VUT \cite{li2023joint} & PoseConv3D & \textbf{J+L} & 64.12 \\
MiGA'23 2nd & NPU-Stanford \cite{huang2023hypergraph} & Hyperformer & \textbf{J} & 63.02 \\
\midrule

MiGA'24 1st & HFUT-VUT \cite{chen2024prototype} & PoseConv3D & \textbf{J+L+R} & 70.25 \\
MiGA'24 2nd & NPU-MUCIS \cite{huang2024multimodal} & Res2Net3D+GCN & \textbf{J+R} & 70.19 \\
\midrule

MiGA'25 1st & HFUT-VUT \cite{gu2025mm} & PoseConv3D+VideoSwinT & \makecell{\textbf{J+L+R+}\\ \textbf{T+F+D}} & 73.21 \\
MiGA'25 2nd & awuniverse \cite{hu2025enhancing} & ViT-Base+GAIE & \textbf{R} & 68.70 \\
MiGA'25 3rd & Lonelysheep \cite{xu2025towards} & PoseConv3D & \textbf{J+L} & 67.01 \\
\midrule

MiGA'26 1st & XInsight Lab & \makecell{PoseConv3D+VideoSwinT\\+SMILE} & \textbf{J+L+R+D} & 74.66 \\
MiGA'26 2nd & Team YUV  & - & - & 74.24 \\
MiGA'26 3rd & Team xd  & - & - & 71.37 \\

\rowcolor[HTML]{f8f9fa}
\textbf{MiGA'26 4th} & \textbf{AIM (Ours)} & \textbf{\makecell{Decoupled ST-CNN \\ + PoseC3D \\ + Swin3D \& R(2+1)D}} & \textbf{J+H+R} & \textbf{68.13} \\
\bottomrule
\end{tabular}%
} % \resizebox 结束

% 使用极其干净的底部脚注方案，不再依赖 threeparttable
\vspace{2mm} % 与表格留出一点呼吸空间
\raggedright % 靠左对齐
\scriptsize $^*$ \textit{Note:} Our proposed framework integrates four complementary branches: pseudo-labeled Decoupled ST-CNN (Skeleton Joint), PoseC3D (3D Heatmaps), and saliency-cropped Swin3D \& R(2+1)D (RGB), fused via temperature-scaled soft voting.
\end{table}

\subsection{Ablation Study}

% 为了验证模块的有效性... 参考别人的论文怎么写的
In Table \ref{tab:ablation}, we report the results of our ablation experiments conducted on several baselines to demonstrate the overall effectiveness of our proposed framework. Furthermore, we illustrate the specific effectiveness of the proposed spatial-temporal encoding, semantic alignment, and cross-modal pseudo-labeling by incrementally integrating these modules.
% 【图表 5：消融实验表格 - 完善图注自解释性】
\begin{table}[htbp]
\centering
\caption{Ablation study of our proposed framework on the iMiGUE Test Set. Modalities: \textbf{J}=Joint, \textbf{H}=Pseudo-Heatmap, \textbf{R}=RGB.}
\label{tab:ablation}
\resizebox{0.95\textwidth}{!}{
\begin{tabular}{lcc}
\toprule
\textbf{Method / Component} & \textbf{Modality} & \textbf{Accuracy (\%)} \\
\midrule
Baseline ST-GCN (35-keypoint) & \textbf{J} & 48.00 \\
Baseline RGB R(2+1)D (without cropping) & \textbf{R} & 46.90 \\
\midrule
High-Resolution Cropped RGB R(2+1)D (w/ TTA) & \textbf{R} & 58.89 \\
PoseC3D (with 68 facial \& body keypoints) & \textbf{H} & 60.58 \\
Skeleton Joint (Decoupled ST-CNN + 68kp + Sem-Loss) & \textbf{J} & 61.70 \\
\textbf{Skeleton Joint (with Cross-Modal Pseudo-Labeling)} & \textbf{J} & \textbf{66.65} \\
\midrule
\rowcolor[HTML]{f8f9fa}
\textbf{Ours} & \textbf{J + H + R} & \textbf{68.13} \\
\bottomrule
\end{tabular}
}
\end{table}

\textbf{Effectiveness of Spatial-Temporal Encoding and Semantic Alignment:} We evaluate the combined effect of upgrading the architecture to Decoupled ST-CNN, increasing keypoints from 35 to 68 (introducing 19 facial landmarks), and applying the Semantic Alignment Loss. As shown in Table\ref{tab:ablation}, this joint optimization boosted the single-modal skeleton accuracy from 48.00\% to 61.70\%, demonstrating the effectiveness of incorporating dense facial cues and semantic regularization for fine-grained gesture recognition.

% 飙升，最强xxx
\textbf{Impact of Cross-Modal Pseudo-Labeling:} The application of cross-modal pseudo-labeling further enhanced the model's performance. By retraining the network from scratch on the combined dataset equipped with pseudo-labeled test samples, the accuracy of the single-modal Skeleton Joint model increased to 66.65\%. This improvement indicates that implicitly aligning the target domain distribution via pseudo-labels effectively mitigates the domain shift issue inherent in cross-subject evaluations.

\textbf{Comparison of Late Fusion Strategies:} We evaluated different late fusion strategies and empirically found that Stacking Classifiers (\eg, LightGBM) are prone to overfitting the limited validation data (777 samples). This introduces extra learnable parameters that degrade generalization on the unseen test set, yielding an accuracy of approximately 56\%. In contrast, the Temperature-Scaled Soft Voting method explicitly calibrates the prediction confidence without introducing additional trainable parameters. Consequently, it exhibits better stability under cross-subject domain shifts and achieves the optimal performance in our final multi-modal ensemble.

\section{Conclusion}

% 我们针对什么问题，提出了什么方法，我们的方法怎么做的，是什么什么结果
In this paper, we presented our solution for the Micro-gesture Classification track of the 4th MiGA-IJCAI Challenge. Our approach is based on a robust multi-modal framework, with the introduction of an iterative cross-modal pseudo-labeling strategy to alleviate the severe domain shift commonly observed in cross-subject evaluation. Additionally, we enhance the model's ability to capture fine-grained human behaviors by integrating 68-keypoint skeleton joint features, 3D pseudo-heatmaps, and high-resolution RGB visual contexts, while addressing the extreme long-tailed distribution via an Orthogonal Semantic Alignment Loss. Our final ensemble model achieved a Top-1 Accuracy of 68.13\% on the test set of the iMiGUE dataset. As a result, our method ranked 4th in this challenge track. 

\begin{credits}
\subsubsection{\ackname}
This work is supported by the National Natural Science Foundation of China (No. 62472393).
\end{credits}

% \section*{Declaration on Generative AI}

% During the manuscript preparation, ChatGPT was used exclusively to assist with grammar correction, spelling checks, and minor language polishing. All outputs were carefully reviewed and revised by the authors, who take full responsibility for the final content of the publication.

% ==========================================
% 引入单独的 bib 库
% ==========================================
\bibliographystyle{splncs04}
\bibliography{ref}

\end{document}